\documentclass[letterpaper]{article} 
\usepackage{aaai2026}  
\usepackage{times}  
\usepackage{helvet}  
\usepackage{courier}  
\usepackage[hyphens]{url}  
\usepackage{graphicx} 
\usepackage{natbib}  
\usepackage{caption} 
\usepackage{algorithm}
\usepackage{algorithmic}

\usepackage{amsmath}
\usepackage{booktabs}
\usepackage{newfloat}
\usepackage{listings}
\DeclareCaptionStyle{ruled}{labelfont=normalfont,labelsep=colon,strut=off} 
\floatstyle{ruled}
\newfloat{listing}{tb}{lst}{}
\floatname{listing}{Listing}
\title{Deadline-Aware, Energy-Efficient Control of Domestic\\Immersion Hot Water Heaters}

\author {
    Muhammad Ibrahim Khan\textsuperscript{\rm 1},
    Bivin Pradeep\textsuperscript{\rm 1},
    James Brusey\textsuperscript{\rm 1}
}
\affiliations {
    \textsuperscript{\rm 1}Centre for Computational Science and Mathematical Modelling, Coventry University\\
    Coventry, United Kingdom\\
    khanm442@coventry.ac.uk, pradeepb2@coventry.ac.uk, aa3172@coventry.ac.uk
}

\begin{document}

\maketitle

\begin{abstract}
Typical domestic Immersion water heater systems are always turned on during the winter, it heats quickly rather than efficiently and ignores predictable demand windows and ambient losses. We study deadline-aware control, where the aim is to reach a target temperature at a specified time while minimising energy. We introduce an efficient Gymnasium environment that models an immersion hot-water heater with first-order thermal losses and discrete on and off actions $\{0, 6000\}$ W applied every $120$ s. Methods include a time-optimal bang-bang baseline, a zero-shot Monte Carlo Tree Search planner, and a Proximal Policy Optimization policy. We report total energy (Wh) under identical physics. Across sweeps of initial temperature $(10–30$ °C), deadline $(30–90$ steps), and target temperature $(40–80$ °C), PPO achieves the most energy-efficient performance at a $60$-step horizon ($2$ h) it uses $3.23$ kWh, versus bang-bang’s $4.37–10.45$ kWh and MCTS’s $4.18–6.46$ kWh, yielding savings of $26$\% at $30$ steps and $69$\% at $90$ steps. In a representative trajectory ($50$ kg, $20$ °C ambient, $60$ °C target), PPO consumes $54$\% less energy than bang-bang and $33$\% less than MCTS. These results show that learned, deadline-aware control reduces energy under identical physics where planners provide partial savings without training, while policies offer near-zero-cost inference once trained.
\end{abstract}


\section{Introduction}

Domestic hot-water heating is a routine service that draws a substantial share of household energy. In practice, demand is clustered at predictable times, such as before work or in the evening. Yet many controllers still drive the element at full power using simple on–off rules \cite{LAKSHMANAN2021, ruelens2017batch, KHURRAM2020}. These rules ignore how much water is in the tank, how quickly heat is lost to the room, and when hot water is actually needed. The result is heating earlier or harder than necessary, higher energy peaks, and missed opportunities to shift load away from carbon-intensive periods. A controller that reasons about volume, ambient conditions, and a specific deadline can heat just in time and just enough, aligning comfort, cost, and emissions goals\cite{kim2024optimalstart, BUECHLER2025, MALTAIS2022}.

Here we study a deadline-aware, energy-minimising control problem for a domestic immersion hot-water heater. The task is to reach a target temperature at a specified time while using as little energy as possible, evaluated against a tolerance band at the deadline. This framing reflects how households plan availability and connects directly to resource management and environmental objectives, without modelling user behaviour patterns.

To make comparisons fair and reusable, we build a lightweight, reproducible simulation environment that captures first-order thermal dynamics with heat loss and exposes a stepwise control interface via the Gymnasium API. Physics, initial conditions, and timing are held constant across controllers so differences in outcome reflect decision making rather than modelling quirks. We evaluate a time-optimal bang-bang baseline that heats as fast as possible, a Monte Carlo Tree Search planner, and a Proximal Policy Optimisation agent, all operating with discrete on and off actions $\{0, 6000\}$ W applied every 120 s. We report total energy in watt-hours (Wh) under identical physics and timing; the example trajectories shown meet the tolerance band at the deadline. We also include small sweeps over target temperature, initial temperature, and deadline (target time step) to show how each controller scales under identical conditions.

A clear pattern emerges. When deadlines are generous or the initial temperature is higher, the energy gap narrows, but bang-bang still trails the other methods. As the deadline tightens or the target temperature increases, anticipation matters where the planner and the learned policy delay or modulate the heating and, in the trajectories shown to reach the target at the deadline with lower energy. PPO in particular tends to avoid overshoot. For deployment, the planner can be competitive on energy but requires online search at each step, whereas a trained policy executes instantly and is easier to embed at scale. Our contributions are: 

\begin{itemize}
    \item A deadline-aware, energy-minimising immersion-heater benchmark with a transparent evaluation protocol under identical physics and timing (discrete on and off actuation, fixed step)
    \item A comparison of bang-bang, MCTS, and PPO on the same environment and
    \item Evidence from small sweeps over target temperature, initial temperature, and deadline that the learned policy (PPO) forms the lower energy envelope while model-based planning (MCTS) provides partial savings over a common baseline, together with a brief discussion of deployment trade-offs between planning and learned policies.
\end{itemize}

\section{Related Works}

Domestic electric water heaters are routinely treated as small thermal stores that can shift heat input without sacrificing service. Work on thermostatically controlled loads and device-level scheduling uses compact first order models to capture standby losses and simple draw dynamics, and then optimises when to charge against energy or tariff objectives while respecting temperature bands at usage times \cite{ruelens2017batch,desomer2017dhw,amasyali2021waterheater}. This view places the decision squarely at the tank: if demand is predictable and the tank stores heat, there is little reason to keep it hot continuously.

Deadline-aware heating is the building-controls version of that intuition. Under the label optimal start, the controller delays heat input so temperature arrives within a band at a specified time, thereby cutting preheat losses \cite{kim2024optimalstart}. In practice this ranges from analytic latest-start rules derived from first order fits to data-driven warm-up predictions. Alongside these, thermostatic on and off control with hysteresis remains the everyday baseline because it is simple and robust. These heuristics set expectations for fixed-power devices and make failure modes visible when deadlines tighten or losses are underestimated.

When a simulatable model is available, model-based planning offers a complementary path. Model predictive control spans linear-quadratic formulations through robust nonlinear variants with explicit constraints and forecasts \cite{drgona2020rser,rawlings2018cace}. Forward search methods such as Monte Carlo Tree Search provide an alternative when one prefers online look-ahead over solving an optimisation problem directly \cite{browne2012mcts}. A consistent theme across these planners is the deployment trade-off: longer horizons and richer branching improve decisions but increase per-step compute and memory, which can exceed the budgets of embedded controllers \cite{putta2013acc}.

Reinforcement learning has been used for HVAC and electric water heaters to trade energy or cost against comfort penalties, from early value-based controllers to policy-gradient agents \cite{wei2017buildsys,ruelens2017batch,rohrer2022hpcontrol}. In well-shaped, stationary tasks, learned policies can approach model-based performance while offering near-zero inference cost at runtime, which makes PPO-style agents attractive once trained. The contrast with planners is practical rather than philosophical: search buys foresight at deployment time, learned policies buy speed.

Finally, benchmark efforts emphasise clear interfaces and transparent metrics so studies are comparable. BOPTEST standardises KPIs and containerised cases for apples-to-apples evaluation, and CityLearn provides Gym-compatible scenarios at district scale \cite{blum2019boptest,citylearn2024}. In the same spirit, we use a minimal Gymnasium environment with identical physics across controllers and report energy at the deadline under a fixed tolerance setting. The focus is a device-level, deadline-aware slice that others can extend with tariffs, emissions signals, or richer disturbances.

Our study adopts the optimal-start intuition from building controls—arrive within a temperature band at a specified time—but focuses on a device-level, deadline-aware slice with a fixed on and off actuation and a first-order tank model. In contrast to cost- or tariff-driven day-scale scheduling common in domestic hot-water studies \cite{ruelens2017batch,desomer2017dhw,amasyali2021waterheater}, we evaluate energy at the deadline under identical physics and timing, without modelling user behaviour, tariffs, or draw events. Compared with model-based planning (MPC or forward search) \cite{drgona2020rser,browne2012mcts}, we place Monte Carlo Tree Search and a learned PPO policy side by side in the same environment to surface the practical trade-off between online search and near-zero inference. Relative to prior RL work on building and water-heater control \cite{wei2017buildsys,ruelens2017batch,rohrer2022hpcontrol}, our contribution is a minimal Gymnasium benchmark that isolates arrival-time control and reports energy with fixed tolerances, enabling clear, method-agnostic comparisons. The scope is deliberately narrow with no time-of-use pricing, emissions signals, or stratified tank models, so that the baseline, planner, and policy can be compared under the same physics. These extensions are natural next steps and remain compatible with the same protocol.

\section{Methodology}

\subsection{Physical system and modelling assumptions}

The simulation environment is a Gymnasium environment which simulates a single water tank with lumped thermal capacity and convective losses to the environment. The water is assumed to be spatially uniform in temperature. Heat is added to the tank by an electric immersion heater with fixed efficiency $\eta$, and heat is lost from the tank via convection. Under these assumptions, the energy balance is
\begin{equation}
m c_p\,\frac{dT}{dt} \;=\; \eta\,P(t)\;-\;hA\,[T(t)-T_a],
\label{eq:ct-balance}
\end{equation}


\begin{table}[t]
\centering
\caption{Environment quantities and fixed settings.}
\label{tab:env_params}
\setlength{\tabcolsep}{3.5pt} 
{\small 
\begin{tabular}{llll}
\toprule
\textbf{Qty} & \textbf{Sym} & \textbf{Val} & \textbf{Unit} \\
\midrule
Temp. & $T$ & -- & $^\circ$C \\
Step temp. & $T_t$ & -- & $^\circ$C \\
Amb. temp. & $T_a$ & 20 & $^\circ$C \\
Spec. heat & $c_p$ & 4184 & J\,kg$^{-1}$\,K$^{-1}$ \\
Heat-loss coeff. & $h$ & 50.0 & W\,${}^\circ$C$^{-1}$ \\
Area & $A$ & 1.5 & m$^2$ \\
Time step & $\Delta t$ & 120 & s \\
Time & $t$ & -- & s \\
Water mass & $m$ & 50 & kg \\
Heater eff. & $\eta$ & 0.95 & -- \\
Elec. power (on) & $P_{\mathrm{elec,on}}$ & 6000 & W \\
Elec. power (off) & $P_{\mathrm{elec,off}}$ & 0 & W \\
\midrule
Traj. $(T_0, T_{\text{tar}}, D)$ & -- & $(20,60,60)$ & $(^\circ\mathrm{C},^\circ\mathrm{C},\text{steps})$ \\
\bottomrule
\end{tabular}
}
\end{table}

\subsection{Discretization and transition function}
Control is applied at a fixed sampling interval $\Delta t=120\,\mathrm{s}$ ($2$ minutes), which means each step time step represents $120$ s. Using forward-Euler integration of~\eqref{eq:ct-balance}:
\begin{equation}
T_{t+1} \;=\; T_t + \frac{\Delta t}{m c_p}\,\big[\eta P_t - hA\,(T_t - T_a)\big].
\label{eq:disc-update}
\end{equation}

With the nominal parameters, the dimensionless cooling factor is
\[
\frac{hA\,\Delta t}{m c_p} \;=\; \frac{75 \cdot 120}{50\cdot 4184} \;\approx\; 0.043,
\]
well within a stable range for explicit integration at $\Delta t=120$\,s.

\subsection{State, observation, and action}
The environment is fully observable there the state space and the action space are exactly the same. The observation space (provided to all controllers) consists of the following scalars:

\begin{equation}
\begin{aligned}
o_t &= \bigl[T_t,\, T_{\mathrm{target}},\, T_a,\, \tau_t\bigr],\\
\tau_t &\equiv T^\star - t .
\end{aligned}
\end{equation}

where $T^\star$ is the designated target time step. The action space is discrete, $\mathcal{A}=\{0,1\}$, mapped to power $\{0,6000\}$\,W.

\subsection{Horizon and termination}
Episodes terminate at the target time, if $\lvert t-T^\star\rvert \le \tau$ (tolerance $\tau\ge 1$ step in evaluation), the transition is terminal and the terminal penalty is applied.

\subsection{Reward shaping and costs}

In order to ensure a balance between energy savings and meeting the time and temperature requirements the reward function is based on the following principles:

\begin{itemize}
    \item In the penultimate step, the cost of energy to improve the final temperature, should be less than the benefit in terms of final reward.
    \item Overall reward for a complete episode should be between either $(1,-1)$, $(0,-1)$ or $(1,0)$.
    \item Final penalty for not reaching target temperature should be uniform.
    \item Penalty for energy used should be uniform.
    \item Reward should only depend on state,action and next state.
    \item Reward function should be as simple as possible.
\end{itemize}

These principles ensure that the devised reward function encourages the model to take heating actions instead of conserving energy throughout the episode and ignoring the temperature requirements.

The reward function can be divided into two parts; the per-step reward, which is a penalty for using energy at that step, and the end of episode reward, which is a penalty based on the temperature difference between the target temperature and the temperature at the end of the episode.  

\begin{equation}
\begin{aligned}
r_t &= -\alpha E_t \\
&\quad +
\begin{cases}
-\beta\,\lvert T_{\mathrm{target}} - T_{t+1}\rvert, & \text{if }\lvert T - T^\star\rvert \le \tau \\[1pt]
0, & \text{otherwise.}
\end{cases}
\end{aligned}
\label{eq:reward}
\end{equation}

with step energy $E_t=P_t\,\Delta t$ (Joules). Constants: $\alpha=1.86\times 10^{-8}$, $\beta=0.03$.

The cost of one extra on-step at the end must be smaller than the benefit of reducing the terminal error by $1^\circ\mathrm{C}$:
\[
\alpha\,E_{\text{step}} \;<\; \beta \cdot 1^\circ\mathrm{C}.
\]

Heater input when “on”:
\[
P_{\text{on}} \;=\; \eta\,P_{\max} \;=\; 0.95 \times 6000 \;=\; 5700~\mathrm{W}.
\]
Step duration:
\[
\Delta t \;=\; 120~\mathrm{s}.
\]
Energy per on-step:
\[
E_{\text{step}} \;=\; P_{\text{on}}\Delta t \;=\; 5700 \times 120 \;=\; 684{,}000~\mathrm{J}.
\]

\subsection*{Selection of $\alpha$}

We target a per-step energy penalty of about $1.3\times 10^{-2}$ so that:
\begin{itemize}
\item Over $60$--$90$ steps, the worst-case all-on energy cost is $\approx 0.76$ to $1.15$, keeping total return within a compact interval near $[-1.2,\,0]$.
\item One extra on-step remains cheaper than a $1^\circ\mathrm{C}$ terminal improvement (see $\beta$ below).
\end{itemize}
Solve $\alpha\,E_{\text{step}} \approx 0.01275 \Rightarrow \alpha \approx 0.01275/684{,}000$, yielding
\[
{\alpha \;=\; 1.86\times 10^{-8}\ \mathrm{J}^{-1}}
\]
and thus $\alpha E_{\text{step}} \approx 0.0128$ per on-step (uniform, state-independent).

\subsection*{Selection of $\beta$}

We require the benefit of improving terminal temperature by $1^\circ\mathrm{C}$ to exceed one on-step cost:
\[
\beta \cdot 1^\circ\mathrm{C} \;>\; \alpha E_{\text{step}} \approx 0.0128.
\]
Choose
\[
{\beta \;=\; 0.03\ /\ ^\circ\mathrm{C}},
\]
which satisfies $\beta(1^\circ\mathrm{C})=0.03>0.01275$. Intuition: in the penultimate step, if one extra on-step can realistically reduce terminal error by $\ 1^\circ\mathrm{C}$, the net gain is $0.03-0.01275\approx 0.01725$; if the expected improvement is small (e.g., $<0.4^\circ\mathrm{C}$), the action is not justified, discouraging gratuitous heating.

\subsection{Experimental factors}
To probe robustness, we vary:
\begin{itemize}
    \item \textbf{Initial temperature:} $T_0\in\{10,15,20,25,30\}^\circ$C
    \item \textbf{Target time step:} $T^\star\in\{30,45,60,75,90\}$
    \item \textbf{Target temperature:} $T_{\mathrm{target}}\in\{40,50,60,70,80\}^\circ$C
\end{itemize}
For each configuration, parameters are set at instantiation; when $T^\star$ increases, $\texttt{max\_steps}$ and the observation bounds are updated in lockstep.

\subsection{Design rationale}
First-order physics with Newtonian losses yields an analytically transparent, computationally light plant suitable for both online planning (MCTS) and policy learning (PPO). The target-time termination emphasizes \emph{when} the target is achieved, not only \emph{whether}. The minimal observation exposes only necessary variables, and binary actuation reflects common immersion-relay hardware while preserving a nontrivial scheduling problem.

We study domestic hot-water heating as a deadline-aware, energy-minimising control problem. A controller must bring the water to a specified target temperature by a given decision time while expending as little energy as possible. To compare alternative strategies on equal footing, we implement a lightweight simulation with first-order thermal losses and a stepwise control loop, so both planning and learning methods act under identical physics and timing.

\subsection{Problem formulation and MDP}

The entire experiment is formulated as a finite-horizon Markov decision process. The state at time $t$ is
\[
s_t = [T_t,\ T_a,\ \tau_t,\ T_{\text{target}},\ M],
\]
The discrete action space,
\[
a_t \in \{0,\ P_{\mathrm{on}}\},\qquad P_{\mathrm{on}} = 6000\ \mathrm{W},
\]
represents off and on at a fixed power level. Episodes truncate at the deadline $t=D$.

A service constraint can be checked at the deadline as
\[
|T_D - T_{\text{target}}| \le \Delta T_{\text{band}},
\]
where $\Delta {T_{\text{band}}} = \pm 1^\circ$C. The results reported in this paper we do not filter runs by success; we report energy for all runs and show a representative trajectory that reaches the target at the deadline.

\subsection{Controllers}

\paragraph{Bang-bang baseline}
A reactive policy applies $P_{\mathrm{on}}$ until the temperature enters the target band, then maintains within the target temperature threshold thereafter. This baseline is time-optimal for reaching the set-point under fixed power but does not explicitly minimise energy.

\paragraph{Monte Carlo Tree Search}

 plans the heater actions over the evaluation horizon by iterating through these canonical four phases: (i) \emph{selection}, where a path is traced from the root using UCB1 with exploration constant $c$ to balance exploitation of high empirical returns and exploration of uncertain branches; (ii) \emph{expansion}, which adds the next unvisited action at the frontier node; (iii) \emph{simulation}, which rolls the exact immersion-heater simulator forward to the horizon to obtain the cumulative reward defined in our study; and (iv) \emph{backup}, which propagates the simulated return to update visit counts and action-value estimates along the path. Because the dynamics are deterministic and the action space is binary, repeated simulations concentrate the rollout budget (25{,}000 per episode) on promising trunks, progressively refining value estimates where it matters most. The final control is the root action with the highest visit count/estimated value, and the resulting action sequence is executed in the same environment used for evaluation, ensuring consistency between planning and deployment.

We use UCB1 for selection,
\[
\arg\max_i \;\frac{Q_i}{n_i} \;+\; c\sqrt{\frac{\ln N}{n_i}},
\]
with \(c=\sqrt{2}\approx 1.414\). In a binary action space (off/on), this value provides balanced exploration: it prevents premature commitment to high-heat branches at shallow depth while allowing rapid concentration on promising trunks once evidence accumulates. Smaller \(c\) risks myopic plans (more overshoot or wasted energy); larger \(c\) spreads simulations too thinly (slower value convergence, higher terminal variance).

 Each plan uses \(25{,}000\) simulations from the root. Budgets below this threshold increased variance; beyond it, improvements were marginal relative to compute.

Achieving \(25{,}000\) rollouts per episode requires a fast, deterministic simulator. We use the same thermodynamic step as evaluation (Newton cooling + heater input), ensuring that MCTS estimates align with execution-time dynamics and that returns reflect the uniform energy and terminal-error costs.

\paragraph{Proximal Policy Optimisation}
We use PPO from Stable-Baselines3 with a multilayer perceptron and a discrete action head. The model was trained under default hyperparameters for 2.5 Million time steps, but the model converged to the optimal policy after 2.1 Million time steps of training on the Immersion water heater simulation environment, with different starting states to ensure generalisibility of the model. 

\subsection{Evaluation protocol}

All controllers are evaluated under identical physics, initial conditions, and timing. The primary metric is total energy $E=\sum_t E_t$ in Wh. We characterise scaling with three one-dimensional sweeps:
\begin{itemize}
\item \textbf{Target temperature:} $T_{\text{target}} \in \{40, 50, 60, 70, 80\}\ ^\circ$C at fixed $T_0$ and $D$.
\item \textbf{Initial temperature:} $T_0 \in \{10, 15, 20, 25, 30\}\ ^\circ$C at fixed $T_{\text{target}}$ and $D$.
\item \textbf{Deadline (target step):} $D \in \{30, 45, 60, 75, 90\}$ at fixed $T_{\text{target}}$ and $T_0$.
\end{itemize}
We additionally show one representative trajectory with $T_0=20^\circ$C, $T_{\text{target}}=60^\circ$C, and $D=60$ steps (each step is $\Delta t=120$ s).

\subsection{Implementation details and reproducibility}

All implementations are in Python. PPO uses Stable-Baselines3. The planner uses the same environment interface. Training and evaluation are performed on CPU.

\section{Results}

In this section we evaluate the performance of PPO and MCTS against the bang-bang approach (baseline). We executed a series of experiments to observe how all of these models perform under different environmental conditions. Each controller must, at 120 s intervals, select a discrete heating action \{0, 6000\} W to drive the water tank from its initial temperature to the specified setpoint at the specified target time step while minimising cumulative energy used.  

For each setting we report the cumulative energy used (Wh), thereby directly comparing these models and evaluating their energy efficiency over various conditions. 

Across all different scenarios PPO consistently outperformed MCTS and baseline in terms of energy efficiency while achieving the temperature requirements. The bang-bang approach was considered as the baseline approach which consumed substantially more energy than other two approaches. The MCTS approach in most cases outperformed the baseline but consistently underperformed against PPO.

\begin{figure}[!t]
\centering
\includegraphics[width=0.9\columnwidth]{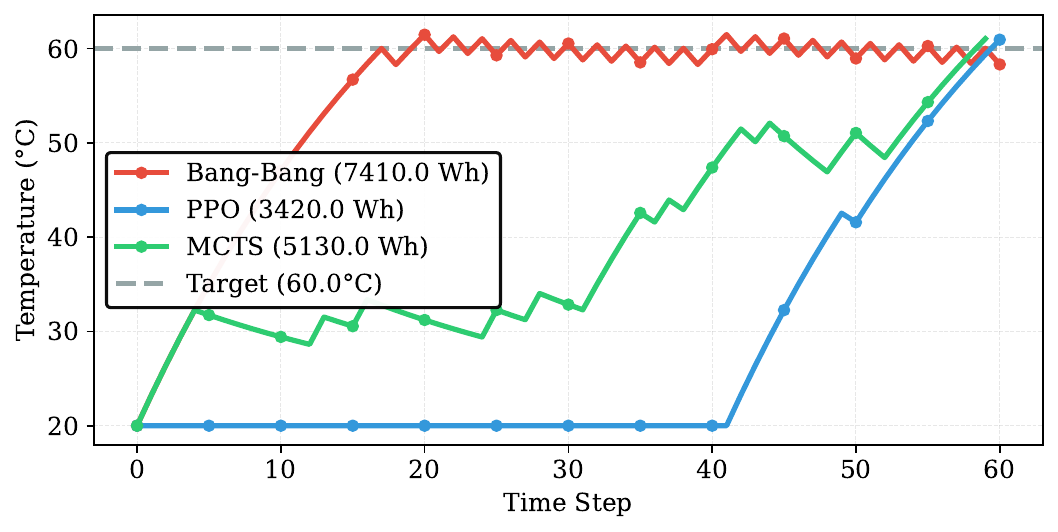}
\caption{Single-episode temperature trajectories under bang–bang, MCTS, and PPO. Setup: $m=50$ kg, $T_a=20^\circ$C, $\Delta t=120$ s, horizon $D=60$ steps, target $60^\circ$C. PPO heats just in time and uses the least energy; MCTS is intermediate; bang–bang is highest.}
\label{fig.trajectory}
\end{figure}

\subsection{PPO Training}

A PPO policy was trained in simulation for 2.5 M environment steps using the default Stable-Baselines3 hyperparameters. To promote generalisation, each episode drew a fresh start state so the policy experienced a broad range of operating conditions rather than a single nominal regime. The learning curve showed a steady increase in episodic return with shrinking variance; performance converged at around 2.1 M steps and subsequently plateaued with only minor stochastic fluctuations, indicating that the learned strategy had stabilised.

\subsection{Zero-Shot MCTS}

As a training-free baseline, MCTS performs online planning with known dynamics and reward at decision time—no data collection, no parameter fitting, and no offline optimisation. In our evaluations, vanilla MCTS serves as a credible zero-shot alternative to PPO: it consistently outperforms a rule-based bang–bang controller on energy efficiency while requiring no training. The trade-off is modestly lower energy efficiency than the converged PPO policy, reflecting the absence of learned priors and amortized optimisation.   

\subsection{Trajectory-level behaviour}

Figure \ref{fig.trajectory} shows a single representative episode with a 60-step horizon and a $60^\circ$C target in the immersion water-heater environment ($m=50$ kg, $T_a=20^\circ$C, $\Delta t=120$ s). PPO follows a delayed heating strategy and settles near the setpoint with the lowest energy, using 54\% less than bang–bang and 33\% less than MCTS. Zero-shot MCTS achieves intermediate savings but exhibits larger initial deviation and less precise terminal adjustment. Bang–bang expends the most energy due to prolonged full-power heating and yields only marginal terminal improvement relative to PPO.

Single-episode trajectories (Temperature vs. Time) reveal the mechanisms behind the aggregate trends. PPO follows the perfect heating strategy with a policy that defers heating until necessary to ensure energy efficiency. Zero-shot MCTS applies targeted bursts that often reduce energy relative to rule-based control but sometimes mis-time terminal adjustments, producing overshoot or residual error. Bang–bang is dominated by prolonged full-power phases and reaches the vicinity of the setpoint late, explaining its large cumulative energy.

\begin{figure}[t]
\centering
\includegraphics[width=0.9\columnwidth]{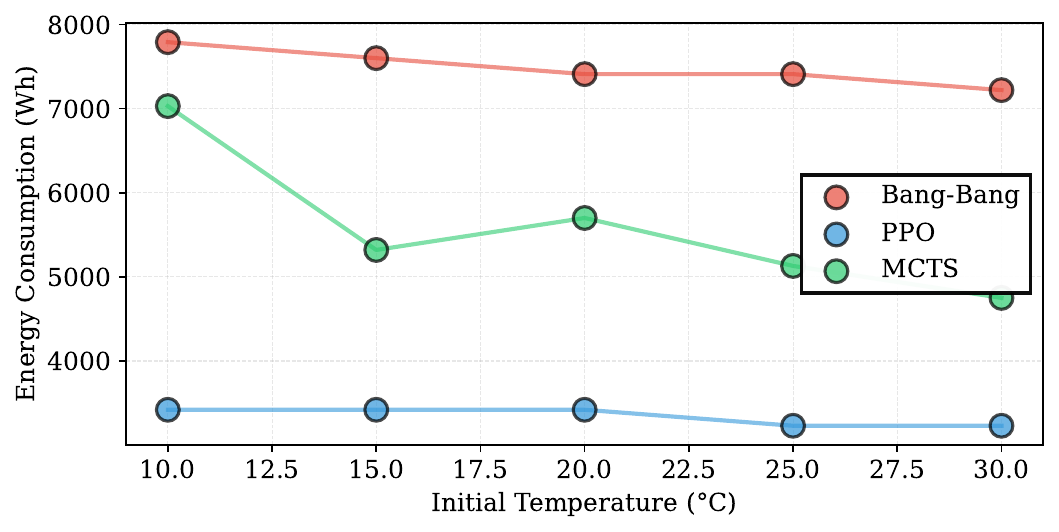} 
\caption{Total energy (Wh) versus initial temperature for three controllers at fixed target $60^\circ$C and horizon $D=60$ steps. PPO forms the lower envelope with low variance; MCTS yields partial savings over bang–bang but is less consistent; bang–bang consumes the most energy across the range.}
\label{fig.initial_temp_scatter}
\end{figure}

\subsection{Initial temperature sensitivity}

Figure \ref{fig.initial_temp_scatter} plots total energy versus initial temperature at fixed $T_{\text{target}}=60^\circ$C and $D=60$ steps: PPO forms the lower envelope with low variance, MCTS is intermediate with higher dispersion, and bang–bang is highest across the range.

\begin{figure}[t]
\centering
\includegraphics[width=0.9\columnwidth]{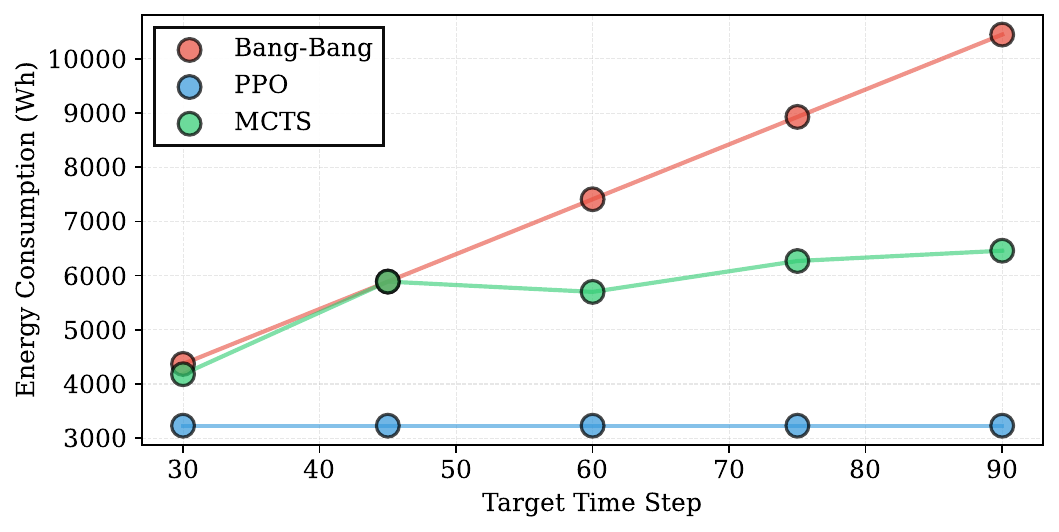}
\caption{Total energy (Wh) versus deadline $D$ (30–90 steps) for three controllers at fixed initial temperature and target $60^\circ$C. PPO remains near-flat across horizons; MCTS is intermediate with some non-monotonicity; bang–bang increases roughly linearly with $D$.}
\label{fig.target_step_scatter}
\end{figure}

\begin{figure}[t]
\centering
\includegraphics[width=0.9\columnwidth]{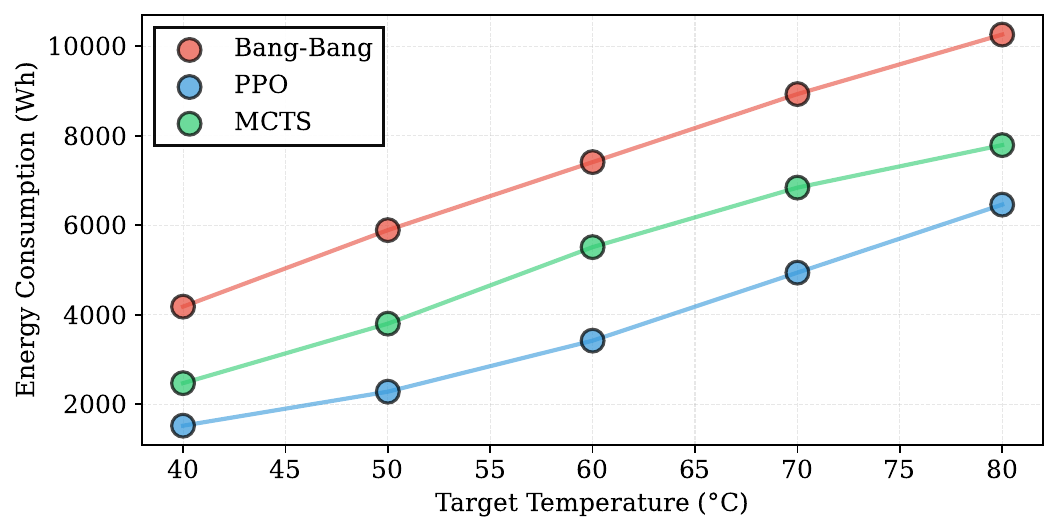}
\caption{Total energy (Wh) versus target temperature (40–80$^\circ$C) at fixed initial temperature and $D=60$ steps. Energy rises for all methods with higher targets; PPO is consistently lowest, MCTS intermediate, bang–bang highest.}

\label{fig.target__temp_scatter}
\end{figure}

Warming the start state reduces the required energy for every controller, but the slope and variance differ dramatically between methods.PPO exhibits weak sensitivity to the initial state and minimal run-to-run dispersion, consistent with a policy that defers heating until necessary, thereby avoiding energy wasted to the higher heat-transfer rates that occur at larger temperature differentials. MCTS yields meaningful savings over bang–bang without any prior training, but its profile is non-monotone in places and the terminal temperature distribution is wider, reflecting search stochasticity and the absence of learned terminal priors. Bang–bang follows the expected inferior energy efficiency with warmer starts and remains dominant in energy use because full-power heating persists until the threshold crossing irrespective of residual horizon.

\subsection{Target time-step (horizon) sensitivity}

Figure \ref{fig.target_step_scatter} shows total energy versus deadline $D\in\{30,45,60,75,90\}$ steps at fixed $T_{\text{target}}=60^\circ$C. PPO remains near-flat at approximately $3230$ Wh across horizons. Bang–bang increases roughly linearly from $4.37$ to $10.45$ kWh (about $1520$ Wh per $+15$ steps, $\approx 101$ Wh/step). MCTS is intermediate (about $4.18$–$6.46$ kWh) with non-monotonic increments consistent with finite search. Relative to bang–bang, PPO’s savings grow with horizon, from about $26\%$ at $30$ steps ($3230$ vs $4370$ Wh) to about $69\%$ at $90$ steps ($3230$ vs $10450$ Wh).

Allowing more time amplifies the differences in time awareness. PPO maintains nearly flat energy across horizons, evidence that the learned policy avoids gratuitous heating when time is abundant and concentrates control effort close to termination. Zero-shot MCTS scales more gently with horizon than bang–bang but is non-monotone due to finite search budgets and horizon-dependent exploration–exploitation trade-offs. Bang–bang energy grows predictably with horizon: longer windows simply extend periods of unnecessary full-power actuation with little improvement in terminal accuracy.

\subsection{Target temperature sensitivity}

Figure \ref{fig.target__temp_scatter} reports total energy versus target temperature $(40–80$$^\circ$C) at fixed initial temperature and $D=60$ steps. Energy rises for all methods with higher targets; PPO is consistently lowest, MCTS intermediate, and bang–bang highest.

Raising the setpoint increases energy required for all controllers. PPO forms the lower envelope with a gradual rise and limited variance. Zero-shot MCTS is intermediate and occasionally shows local irregularities, symptomatic of myopic rollouts that do not fully internalize end-effects at higher setpoints. Bang–bang exhibits the steepest growth, driven by its binary actuation and insensitivity to remaining time.

\subsection{Summary of Quantitative Results}

PPO consistently occupies the Pareto-efficient frontier: lowest energy at near-setpoint terminal temperatures and the tightest dispersion across all sweeps. Zero-shot MCTS is strictly better than bang–bang on energy in most settings and approaches PPO in some regimes, but it exhibits higher terminal-temperature variance and occasional overshoot/undershoot. Bang–bang achieves acceptable terminal temperatures but pays for them with systematically higher energy, particularly as the permitted horizon or the setpoint increases.

Throughout these experiments a consistent theme can be observed for all three models. The bang-bang controller heats at full power from the very start and once it reaches the target temperature, it oscillates within the target temperature range until the target timestep is reached. This is the least energy efficient approach and it loses significant amount of energy while  actively trying to maintain the temperature. MCTS on the other hand is a good mix of performance and no training, as it consistently outperformed bang-bang approach. MCTS controller was unable to outperform PPO, because the MCTS applied targeted bursts of heating but it can overshoot due to the randomness introduced in the rollout. PPO learnt the optimal policy, where it would follow a delayed heating strategy to avoid losing heat to environment thereby conserving energy. The controller consistently used the least amount of episode energy, underscoring the significance of prior training.

\section{Conclusion}

We studied deadline-aware control for a domestic immersion hot-water heater under identical physics and timing, using an efficient Gymnasium environment with discrete on and off actions applied every 120 s. Across three families of experiments: sweeps over initial temperature, target time step (deadline), and target temperature, and a representative trajectory, a clear pattern emerged. The learned policy (PPO) consistently formed the lower energy envelope, the zero-shot planner (MCTS) provided partial savings without training, and the time-optimal bang-bang baseline consumed the most energy; therefore, it was least energy efficient approach. In the $60$-step, $60$\,$^\circ$C case, PPO reduced total energy substantially relative to bang-bang and MCTS, and the trajectories shown reach the target at the deadline with lower energy.

These findings reinforce a simple message that when the task is to arrive on time while using less energy, anticipation matters. Controllers that delay or modulate heating near the deadline avoid unnecessary losses, while binary full-power behaviour pays an increasing penalty as target temperature rises or the available time changes. The planner versus policy trade-off is practical where MCTS offers training-free improvements but incurs online search at every step, whereas a trained PPO policy executes instantly and is easier to embed at scale.

Future work will extend this to continuing on-demand control with periodic truncation and incorporate time-varying tariffs and richer actuation to assess cost, emissions, and peak power alongside energy.


\bibliography{aaai2026}


\end{document}